\documentclass[10pt,twocolumn,letterpaper]{article}

\usepackage{iccv}
\usepackage{times}
\usepackage{epsfig}
\usepackage{graphicx}
\usepackage{amsmath}
\usepackage{amssymb}
\usepackage{url}
\usepackage{array,makecell}
\usepackage{authblk}
\usepackage{lipsum}

\newcommand\blfootnote[1]{%
	\begingroup
	\renewcommand\thefootnote{}\footnote{#1}%
	\addtocounter{footnote}{-1}%
	\endgroup
}

\usepackage[breaklinks=true,bookmarks=false]{hyperref}

\iccvfinalcopy 

\def\iccvPaperID{****} 

\ificcvfinal\fi

\begin{document}

\title{Multi-attention Networks for Temporal Localization of Video-level Labels}


\author[1,*]{Lijun Zhang}
\author[1,*]{Srinath Nizampatnam}
\author[1,*]{Ahana Gangopadhyay}
\author[2,*]{Marcos V. Conde}
\affil[1]{Washington University in St. Louis, St. Louis, United States}
\affil[2]{University of Valladolid, Valladolid, Spain}
\affil[*]{All authors contributed equally to this work} 
\affil[ ]{\tt \small {\{lijunzhang, srinath, ahana\}@wustl.edu, marcosventura.conde@alumnos.uva.es}}

\maketitle

\begin{abstract}
   Temporal localization remains an important challenge in video understanding. In this work, we present our solution to the 3rd YouTube-8M Video Understanding Challenge organized by Google Research. Participants were required to build a segment-level classifier using a large-scale training dataset with noisy video-level labels and a relatively small-scale validation dataset with accurate segment-level labels. We formulated the problem as a multiple instance multi-label learning and developed an attention-based mechanism to selectively emphasize the important frames by attention weights. The model performance is further improved by constructing multiple sets of attention networks. We further fine-tuned the model using the segment-level dataset. Our final model consists of an ensemble of attention/multi-attention networks, deep bag of frames models, recurrent neural networks and convolutional neural networks. It ranked 13th on the private leaderboard and stands out for its efficient usage of resources.
   \blfootnote{This work was presented at the 3rd Workshop on YouTube-8M Large-Scale Video Understanding, at the International Conference on Computer Vision (ICCV 2019) in Seoul, Korea.}
\end{abstract}

\section{Introduction}

With the fast development of digital recording devices and online video sharing platforms, the number of videos available is increasing exponentially, making video understanding a challenging problem in computer vision. As a first step towards video understanding, a significant amount of work has been dedicated to video classification. However, the video understanding problem goes beyond a standard classification problem. Temporally localizing the presences of objects/actions can help us to identify relevant moments in a video and thus better understand its content. Moreover, a video can contain a number of topics that are not always characterized by every time segment within the video. Hence, a better temporal localization algorithm can enable applications such as improved video search (search within a video), highlights extraction, etc. To accelerate the research of temporal localization, Google AI recently released the YouTube-8M Segment Dataset. In this work, we focus on a segment-level classification task presented in the third YouTube-8M Challenge using this segment-level dataset.
\par 
Previous YouTube-8M challenges focused on developing models for video-level predictions. Standard deep neural networks like convolutional neural networks (CNNs) \cite{karpathy2014large,wang2017monkeytyping} and recurrent neural networks (RNNs) \cite{li2017temporal,ostyakov2018label} have been used for video-level classification and both have achieved state-of-the-art results. Pooling via clustering schemes, such as Vector of Locally aggregated Description (VLAD) \cite{jegou2010aggregating,miech2017learnable} and deep bag-of-frames (DBoF) \cite{abu2016youtube,sivic2003video}, has also been widely used among the top competitors. However, these frame-level classifiers are designed to classify video-level labels and cannot necessarily perform segment-level predictions well. Different temporal action localization networks have also been proposed to solve the temporal localization problem. One popular structure is a two-stage, proposal-plus-classification framework \cite{chao2018rethinking}. But to utilize large video-labeled training sets, such a model cannot be directly applied. 
\par 
To better leverage the large training dataset which only has noisy video-level labels together with a comparatively smaller segment-level validation dataset, we propose to utilize a multi-instance learning (MIL) model \cite{maron1998framework,ilse2018attention} to simultaneously temporally localize and classify the target segments. The core idea is to use multiple attention weights to emphasize critical frames from different high-level topics in the video.  The proposed model performed better than both standard neural networks and pooling via clustering methods using our training procedure. Before discussing the models for the task in this challenge, we will give a brief overview of the dataset and the unique challenges it poses.

\subsection{YouTube-8M Segment Dataset}
The YouTube-8M Segment Dataset is an extension of the previous YouTube-8M Dataset \cite{abu2016youtube,lee20182nd} which includes human-verified segment-level labels. The previous YouTube-8M dataset contains 6 million high-quality video samples from YouTube, which were split into 3 partitions: training, validation and test sets, following approximately $70\%$, $20\%$ and $10\%$ split. The video samples were encoded as a hidden representation produced by Inception-v3 \cite{szegedy2016rethinking} pretrained on the ImageNet dataset \cite{ioffe2015batch} for both audio and video features taken at a rate of 1 Hz. This dataset only contains video-level annotations with 3862 class labels and an average of 3 labels per video.
\par 
In the YouTube-8M Segment Dataset, multiple 5-second segments are sampled based on classifier scores to encourage both positive and negative within a video. Then each segment is labeled by human raters from a subset of original vocabulary, excluding entities that are not temporally localizable. In total, ~237K segments covering 1000 categories are labeled. These video segment labels provide a valuable resource for temporal localization. In the 3rd YouTube-8M video understanding challenge, we are required to predict segment-level labels in the test set. Submissions are evaluated using the Mean Average Precision @$K_s$ ($\text{MAP}$@$K_s$), where $K_s=100,000$. For each entity, the $\text{MAP}$@$K_s$ score is calculated as
\begin{equation}
\text{MAP}@K_s = \frac{1}{C} \sum_{c=1}^C \frac{\sum\limits_{k=1}^{K_s} P(k)rel(k)}{N_c}
\end{equation}
where $C$ is the number of Classes, $P(k)$ is the precision at cutoff $k$, $K_s$ is the number of segments predicted per class, $rel(k)$ is an indicator function equaling $1$ if the item at rank $k$ is a relevant (correct) class, or zero otherwise, and $N_c$ is the number of positively-labeled segments for each class.
\par 
The paper is organized as follows. In section 2, we review the general model architectures we used, along with some related work. Details of our models are introduced in section 3. Section 4 presents the evaluation of all models, and section 5 concludes the paper.

\section{Related work}
Most of the models used in previous challenges consist of two components, frame-level pooling and a video-level classifier. The first component pools frame-level features over time and the second component classifies the pooled features. For the pooling component, the DBoF \cite{abu2016youtube} model typically projects the features from a fixed number of randomly selected frames in a video into a higher dimensional vector and pools across frames in that space to create a video-level representation. NetVLAD \cite{arandjelovic2016netvlad} determines differentiable soft assignments of frames to different clusters and uses the concatenation of residuals for each visual word as the final representation. RNNs extract hidden representations of frames over time and use the final output state as a representation of the video. CNNs use convolution kernels to extract higher level representations of features and implements max-pooling across time. For the classifier component, typically a logistic layer or a Mixture of Experts (MoE) \cite{jordan1994hierarchical} are used to obtain the final predictions. Other tricks, including context gating \cite{miech2017learnable}, data distillation \cite{wang2017monkeytyping,skalic2018building} and exponential weight averaging \cite{skalic2018building} have also been reported to boost the final prediction outcome. Our solution too, in the same vein, consists of two components. However, in our model, we developed an attention/multi-attention-based mechanism to pool the frame-level features.


\section{Models}
In this section, we will discuss how we formulated the problem and how we used a multi-attention model to address the challenge of temporal localization. 

\begin{figure*}
	\begin{center}
		\includegraphics[page=1,width=\textwidth]{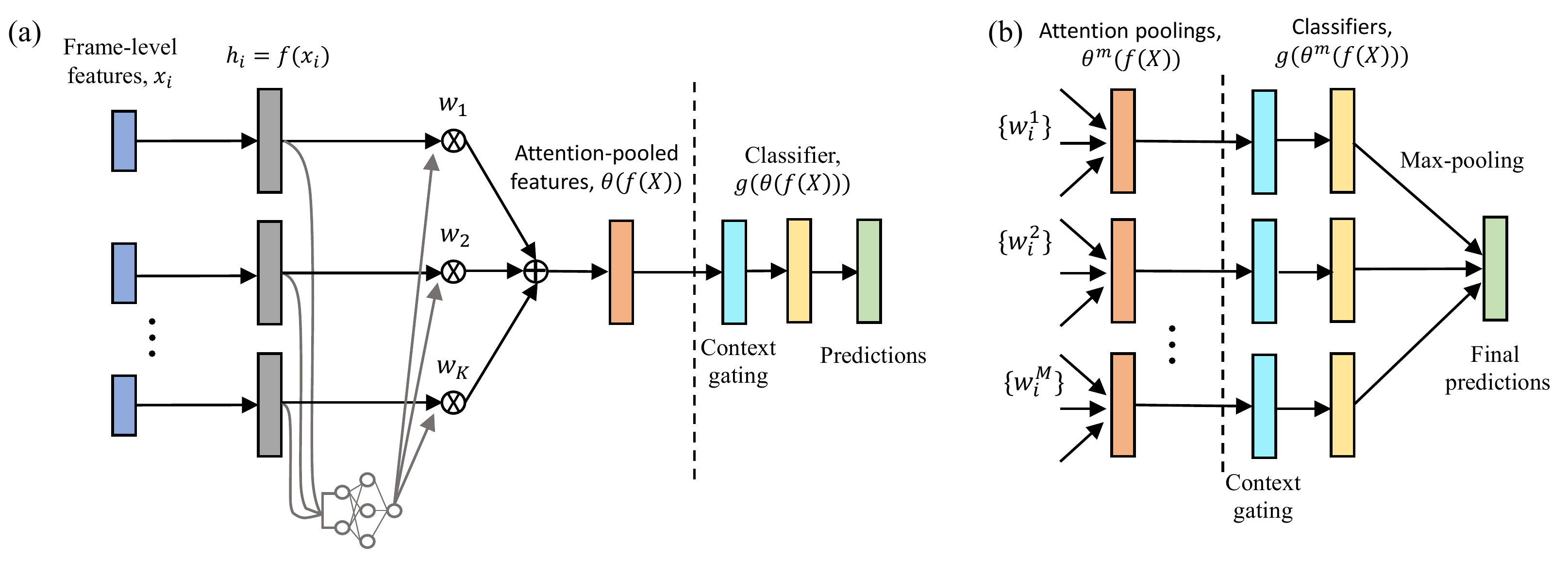}
	\end{center}
	\caption{(a) A schematic showing attention-based pooling for temporal localization. Frames are pooled using an attention network with frame-level features. (b) An extension of the model shown in (a), where multiple attention networks are used.}
	\label{fig:attn}
\end{figure*}

\subsection{Multiple instance multi-label learning}
Multiple instance learning (MIL) deals with problems with incomplete knowledge of labels in the training set. In the case of MIL, there is a bag of instances, $X=\{x_1,x_2,...,x_K\}$, where $K$ represents the total number of instances in the bag. However, we only have access to the label $y$ associated with the bag instead of the labels of individual instances. In the YouTube-8M frame-level video dataset, each frame at one timepoint $x_i$ can be considered as one instance in the bag of frames. Furthermore, each video (bag of frames) is annotated with multiple class labels $y \in [0,1]^n$ (n=3862 in the training set and $n=1000$ in the validation and testing set).
\par 
A generic MIL model can be described as
\begin{equation}
S(X) = g \Bigg(\theta \Big(f(X) \Big) \Bigg).
\end{equation}
The choice of the functions $g,\theta,f$ determines the specific model to predict label probability of the bag. In the YouTube-8M competition, a majority of the commonly used models can be categorized as embedding-based MIL methods. In the embedding-based methods, the transformation $f$ maps each instance to a low-dimensional feature space. Then individual instances are aggregated by the MIL pooling $\theta$ and finally classified by the model $g$. Taking two baseline models as examples, each component can be understood as follows
\begin{itemize}
\item Frame-level logistic model: The pretrained Inception-v3 network first maps the original image and audio at each time point into a frame-level feature. Then frame-level features are pooled across the bag by the mean operator
\begin{equation}
\theta \Big(f(X) \Big) = \frac{1}{K} \sum_{i=1}^K f(x_i).
\end{equation}
Finally, the pooled features are classified by a logistic regression model.
\item Frame-level DBoF model: The frame-level features extracted from the Inception-v3 network are projected onto a higher dimensional space. Then the max pooling was used to perform the aggregation
\begin{equation}
\theta \Big(f(X) \Big) = \max\limits_{i=1,...,K} f(x_i).
\end{equation}
Other MIL pooling methods can be used to replace the mean or max operator, such as log-sum-exp \cite{ramon2000multi}, noisy-and \cite{kraus2016classifying}, etc. But to better emphasize the frames that contribute most to the final prediction result, we propose a learnable weighted average of frames as the pooling method.
\end{itemize}


\subsection{Gated attention network}
In the training dataset, each bag contains all the frames from a video. In the validation/test dataset, the model only needs to predict the labels of 5-frames segments. To bridge the gap between the regular training set and the validation/testing set, it is particularly important that the pooling method can emphasize critical frames or segments of a video. We formulate the pooling as a weighted average of frames
\begin{equation}
\theta \Big(f(X) \Big) = \sum_{i=1}^K w_i h_i
\end{equation}
where, $h_i=f(x_i)$ are the frame-level features after initial mapping. The attention weights $w_i$ are determined by a neural network
\begin{equation}
w_i = \frac{\exp \Big(a^T tanh(Vh_i^T) \Big)}{\sum\limits_{j=1}^K \exp \Big(a^T tanh(Vh_j^T) \Big) }
\end{equation}
where $V∈ \in \mathcal{R}^{L \times D}$ and $a \in \mathcal{R}^{L \times 1}$ are parameters of the neural network. The softmax function is used to enforce the sum of weights to 1 such that it is invariant to the length of a video. Furthermore, an additional gating mechanism using sigmoid function \cite{dauphin2017language} along with $tanh⁡(∙)$ nonlinearity is used to help the attention network learn complex non-linear relationships among instances \cite{ilse2018attention}
\begin{equation}
w_i = \frac{\exp \Bigg(a^T \Big(tanh(Vh_i^T) \bigodot \sigma(Uh_i^T) \Big) \Bigg)}{\sum\limits_{j=1}^K \exp \Bigg(a^T \Big(tanh(Vh_j^T) \bigodot \sigma(Uh_j^T) \Big) \Bigg) }
\end{equation}
where $U \in \mathcal{R}^{L \times D}$ are parameters, $\bigodot$ is an elementwise multiplication and $\sigma(.)$ is sigmoid non-linearity.
Once frame-level features are pooled by weighted averaging, a video-level classifier model is used to obtain final prediction probabilities, as shown in Figure 1 (a).
\par  
We used attention weights $w_i$ to select the important frames in a video. Such attention mechanism can also be considered as a temporal localization scheme, in which frames or segments with more discriminant information will be emphasized. We further experimented with a sparsemax function \cite{martins2016softmax} instead of softmax, to have a more selective and compact focus on frames. The sparsemax model is also included in the final ensemble of models.

\subsection{Multi-attention network}
\begin{figure*}
\begin{center}
\includegraphics[page=1,width=0.95\textwidth]{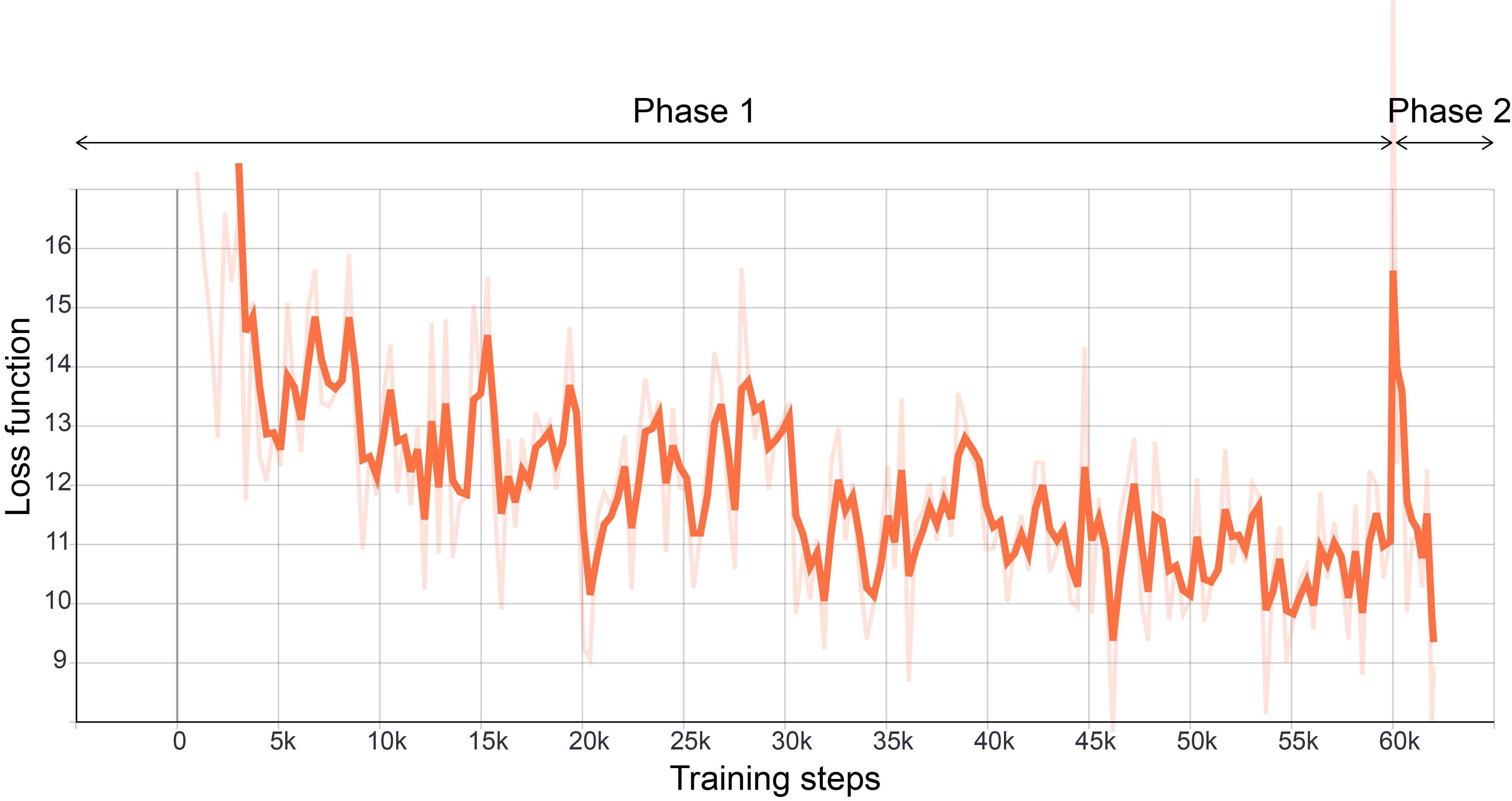}
\end{center}
\caption{The loss function for an attention network over two phases of training. Loss function was computed using cross entropy over labels with more weights assigned to the 1000 classes that are present in the segment-level dataset. Evolution of the loss function during fine-tuning step (Phase 2) is also shown.}
\label{fig:loss}
\end{figure*}
Attention networks need to detect the presence of any frames that contain target information. As individual videos may contain multiple labels, it may be difficult for a single attention network to detect all the topics present.  Therefore, we extended our attention network to a multi-attention network, where we used multiple sets of parameters for the attention network $\{a^m,V^m,U^m \}, \enskip m=1,...,M$, to obtain different sets of attention weights $W^m$. Frame-level features were subsequently pooled by different attention weights
\begin{equation}
\theta^m \Big(f(X) \Big) = \sum_{i=1}^K w_i^m h_i
\end{equation}
Each pooled feature $\theta^m (f(X))$ was then fed into the video-level classifier separately
\begin{equation}
S^m(X) = g \Bigg(\theta^m \Big(f(X) \Big) \Bigg)
\end{equation}
Finally, as shown in Figure 1 (b), the prediction outputs were pooled to obtain the final prediction result 
\begin{equation}
S(X) = \max\limits_{m=1,...,M} S^m(X).
\end{equation}
In our final submission, we used 8 or 16 sets of parameters ($M=8$ or $16$) to select frames differently. The number roughly matched the top-level topic in the vocabulary. We found that more sets of parameters than these led to worse results.

\subsection{Other models used }
In addition to the above-mentioned attention networks, we also adapted gated attention mechanisms to train RNN and DBoF models. For the RNN models, the final output state contains critical information about the video. But there is a possibility that the actions to be classified get masked by other non-relevant actions. This can be problematic particularly when we train using very long videos. To overcome this problem, we performed pooling across all the hidden states using gated-attention mechanism. In the DBoF models, we also experimented using attention-based pooling to aggregate the up-projected features. For all the models, the video and audio features were transformed and pooled separately. The features were first fed into a context gating layer before they were sent to the classifier component.

\section{Experiments}

\subsection{Training procedure}
To better utilize the large frame-level training set and the segment-annotated validation set, we divided our training procedure into two phases. In phase 1, we trained the model on the 1.4 TB regular training set. For each video, we sampled a subset of frames and used attention network to pool across those them. We tried two ways of sampling, random sampling with replacement and sampling one frame every five frames. We found different models favor different sampling schemes. The first and the last 15 frames were excluded as they may only contain the title, ending and credit frames of the video. In phase 2, we fine-tuned the model pre-trained on the regular training set using the validation set with segment labels. In both phases, cross entropy was used as the loss function. As the test set contains only 1000 class labels, more weights were assigned to those classes during training. Each model was trained using the Adam optimizer with a batch size of 128. Models were trained for about 60K steps in phase 1 and around 2K to 3K steps in phase 2. A plot of the loss function over time during the two phases of training is shown in Figure 2. As the number of segments with human-annotated labels are comparatively smaller than the regular training set, more training steps in Phase 2 would make the model overfit to the segment-level dataset and lead to worse prediction outcomes. We also found that more training steps in phase 1 led to worse results, also due to over-fitting. All the training jobs were done in Google Cloud Platform using a single P100 GPU (for attention/multi-attention models, this took around 6 hours in phase 1 and 20 minutes in phase 2).

\subsection{Results}
\begin{table}
\begin{center}
\begin{tabular}{|c|c|}
\hline
\textbf{Model} & \textbf{MAP@100,000} \\
\hline
\makecell{Attention 1 \\ (120 samples, Sparsemax, MoE)} & 0.769 \\
\hline
\makecell{Attention 2 \\ (subsampling, Softmax, MoE)} & 0.768 \\
\hline
\makecell{Attention 3 \\ (120 samples, Softmax, Logistic)} & 0.768 \\
\hline
\makecell{Multi-attention 1 \\ (8 sets, Logistic)} & 0.771 \\
\hline
\makecell{\textbf{Multi-attention 2} \\ \textbf{(8 sets, MoE)}} & \textbf{0.772} \\
\hline
\makecell{\textbf{Multi-attention 3} \\ \textbf{(16 sets, MoE)}} & \textbf{0.772} \\
\hline
\end{tabular}
\end{center}
\caption{Performance of Attention/Multi-attention models.}
\end{table}
Our implementation is based on the TensorFlow starter code\footnote{\url{http://github.com/google/youtube-8m}} and 2nd year's winning solution\footnote{\url{https://github.com/miha-skalic/youtube8mchallenge}}. The main results of our attention/multi-attention models are summarized in Table 1. The Mean Average Precision @K(MAP@K), where K=100,000 was used to evaluate the models. All the MAP@100,000 scores are obtained from the public leaderboard\footnote{\url{  https://www.kaggle.com/c/youtube8m-2019/leaderboard}}.
\par 
We experimented with different variations of attention/multi-attention models. Variations in the attention networks include sampling schemes (randomly sampling 120 frames with replacement or sample 1 frame every 5 frames), output layer non-linearity of the attention network (Softmax or Sparsemax) and the video-level classifier used (logistic layer or MoE). Variations in the multi-attention networks include output layer non-linearity (Softmax or Sparsemax) and the number of parameter sets used (either 8 or 16). All the models were first trained on the frame-level video training set and fine-tuned on the segment-level validation dataset. The finetuning procedure generally improved the scores by around $0.05$. From the evaluation results, we found that a more complex video-level classifier (the size of models with MoE classifier is around 140 MB and the size of models with logistic classifier is around 25 MB) will not necessarily improve the prediction performance. Such effects may be due to the limited size of the ground truth segment-level dataset. Moreover, adding multiple attention networks would only increase the overall model size by 10 MB, which makes our models very resource efficient. In our models, the gating mechanism was mainly used in two places: gated attention network and context gating after frame-level pooling. The first gating mechanism is designed to help the network learn complex relations across frames. The context gating is used to learn non-linear interactions among activations of features. Both gating mechanisms improved the model performance.
\par 
Other models used in the final ensemble were also evaluated using the same procedures and the public leaderboard scores are reported in Table 2. 
\begin{table}
\begin{center}
\begin{tabular}{|c|c|}
\hline
\textbf{Model} & \textbf{MAP@100,000} \\
\hline
CNN1 & 0.757 \\
\hline
CNN2 & 0.755 \\
\hline
DBoF1 & 0.763 \\
\hline
DBoF2 & 0.757 \\
\hline
NetVLAD  & 0.753 \\
\hline
GRU & 0.758 \\
\hline
\end{tabular}
\end{center}
\caption{Performance of other models.}
\end{table}
\begin{figure}[t]
\begin{center}
\includegraphics[width=0.95\linewidth]{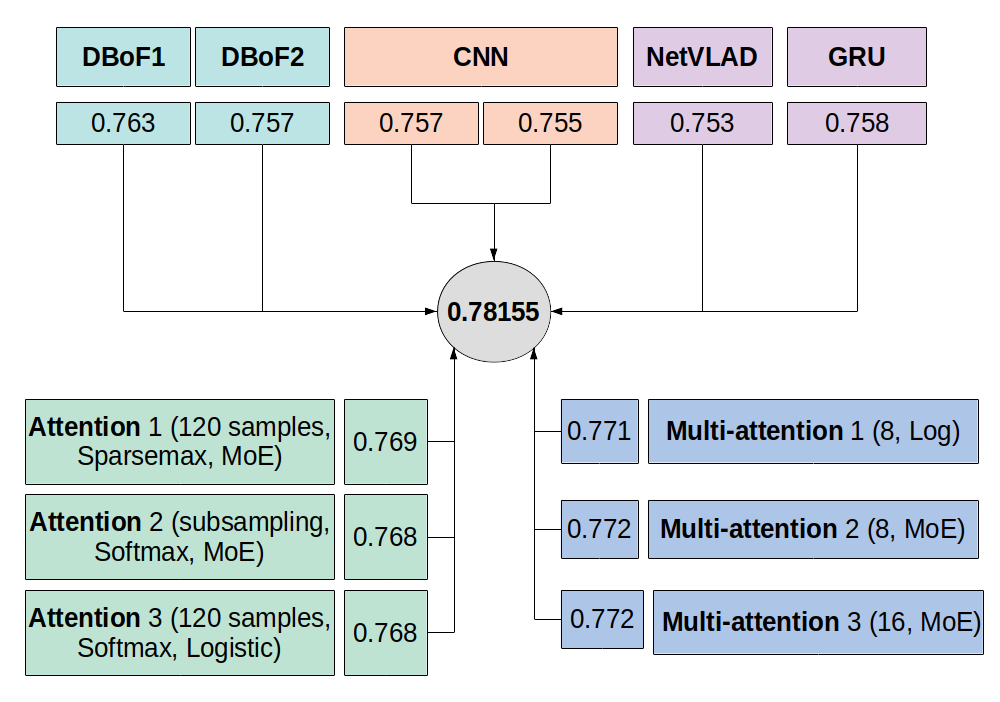}
\end{center}
\caption{Ensemble of all the models trained and their respective MAP@100,000 scores.}
\label{fig:long}
\label{fig:onecol}
\end{figure}
The two CNN models used differ in their sampling schemes. The first CNN model subsamples frames (1 frame every 5 frames) whereas the second randomly samples a sequence of frames.  Out of the two DBoF models, the first uses max-pooling and the second uses attention-based pooling. For recurrent neural network models, we found that using Gated Recurrent Units (GRU) results in a slightly better performance compared to Long Short-term Memory (LSTM) units. Additionally, we tried using bi-directional units instead of uni-directional units. However, the model performance deteriorated using bi-directional units. 
\par 
Different models favor different sampling schemes in the first phase of training. CNN models give a better performance when the video is subsampled every 5 frames. Sampling a sequence of frames with a length of 150 works best for the RNN model. For DBoF, a much smaller sampling size, 30 frames, is found to be optimal. For attention/multi-attention models, sampling $90-150$ frames gave similar results. The final ensemble consists of a weighted summation of the 12 models reported before, as shown in Figure 3.

\section{Conclusion}
In this work, we presented a multi-attention model to address the challenge of temporal localization for video understanding. We demonstrated the effectiveness of  attention-based mechanism to identify important frames over time. And we found that using different attention networks to detect frames from different topics improved the prediction outcome. Such method narrows down the gap between the video-level labeled training dataset and the segment-level labeled validation dataset. Code used in this challenge, as well as the full model architectures and learning parameters, are available at \url{https://github.com/mv-lab/youtube8m-19}. 
\par 
There are multiple potential directions to improve our current model. For attention models, we only used video-level labels or segment-level labels to train the parameters. However, in the validation dataset, start times and end times for the segments are also provided. It will be beneficial to use the start time information as another supervisor. We can add another loss related to segment timing information and the weights put to that segment by the attention network to the loss function. This may enable the attention network to predict the important segments more accurately. In phase 2 of training, we only used 5 frames within the segment as our input to the models. Including the context of that segment will also help the models better understand and classify segment entities. Finally, as the validation dataset is comparatively small, data-augmentation may benefit the model. Besides manipulation of the validation dataset itself, such as producing “virtual” segments by linear combinations of existing segment samples, a pseudo-labeling procedure may also help. A typical pseudo-labeling procedure will choose the top scorer segments in the test set as new training samples for the models. All these methods will potentially improve the performance of our models.

\section{Acknowledgements}
We would like to thank Kaggle and the Google team for hosting the YouTube-8M video understanding challenge and for providing the starter YouTube-8M TensorFlow code. We would also like to thank the previous participants for sharing their solutions.

{\small
\bibliographystyle{ieee_fullname}
\bibliography{egbib}
}

\end{document}